\documentclass[runningheads]{llncs}
\usepackage{graphicx}
\usepackage{comment}
\usepackage{subfig}
\usepackage{wrapfig}

\begin{document}
\title{Guide-Guard: Off-Target Predicting in CRISPR Applications}
\titlerunning{Guide-Guard}
%

\author{Joseph Bingham\inst{1} \and
Netanel Arussy\inst{1} \and
Saman Zonouz\inst{1}}
\authorrunning{J. Bingham et al.}
%
\institute{Rutgers University, New Bruswick, NJ 08901, USA \\
\email{\{joseph.bingham, netanel.arussy, saman.zonouz\}@rutgers.edu}}

\maketitle              
\begin{abstract}
With the introduction of cyber-physical genome sequencing and editing technologies, such as CRISPR, researchers can more easily access tools to investigate and create remedies for a variety of topics in genetics and health science (e.g. agriculture and medicine). As the field advances and grows, new concerns present themselves in the ability to predict the off-target behavior. In this work, we explore the underlying biological and chemical model from a data driven perspective. Additionally, we present a machine learning based solution named \textit{Guide-Guard} to predict the behavior of the system given a gRNA in the CRISPR gene-editing process with 84\% accuracy. This solution is able to be trained on multiple different genes at the same time while retaining accuracy.


\end{abstract}
\section{Introduction}

The newly found simplicity and versatility of gene editing systems such as the Clustered Regularly Interspaced Short Palindromic Repeats (CRISPR)~\cite{blackburn2013crispr} have spawned interdisciplinary research that explores the power and applicability of sequencing, modifying and editing genomes of model organisms across different domains of life. These promising results have not only prompted the discussion of clinical trials in humans, but have already been tested on humans. As the number of viable genes therapies increases, so too does the need to be a way for practitioners to be able to understand the underlying biological and chemical models that determine the safety of the treatment.

CRISPR is the defense mechanisms of prokayotic organisms, such as some bacteria, against infecting viruses. It does this by capturing a sub-sequence of the infecting viruses' DNA and insert it into the cell's DNA in a particular pattern to create segments. These patterns are then used as guides, for future CRISPR to seek the sequence and cut sequences that contain them. In this capacity, it is the organisms highly adaptive immune system~\cite{RATH2015119}.

CRISPR requires a guide RNA (gRNA)~\cite{guide_seq} sequence, which determines where it will cut the given transcriptome~\cite{transcriptome}. In the prokayotic organism, this cut in the genome would kill the would be infectious virus. In modern gene therapies, the guide sequence would be selected to be a part of the gene that is problematic. This would either kill the sequence, or induce a closure of the shortened gene sequence or to include a new sequence. 

As these technologies become more understood, the medical community is also expanding which deceases they are hoping to cure using CRISPR. The main challenge in the desire to ensure the safety of a given gRNA. This is currently done by testing the guide on living cells~\cite{testing}, however this is time consuming and expensive. 

As these cyber-physical technologies are increasingly trending towards relying on computers for automation and verification of the gene editing process, so too does the need for automatic verification and validations. This has spawned a new field of research referred to as \textit{cyberbiosecurity}. The devastating effects of cyberbiosecurity threats, or unverified gRNA's being used are obvious given the importance of certain applications, e.g., human cells, diagnostics and crops. Unlike other domains, the consequences of an unverified guide directly impact a human's way of living and even lives themselves.

\begin{wrapfigure}{r}{.3\textwidth}
  \centering
  \includegraphics[height=0.3\textwidth]{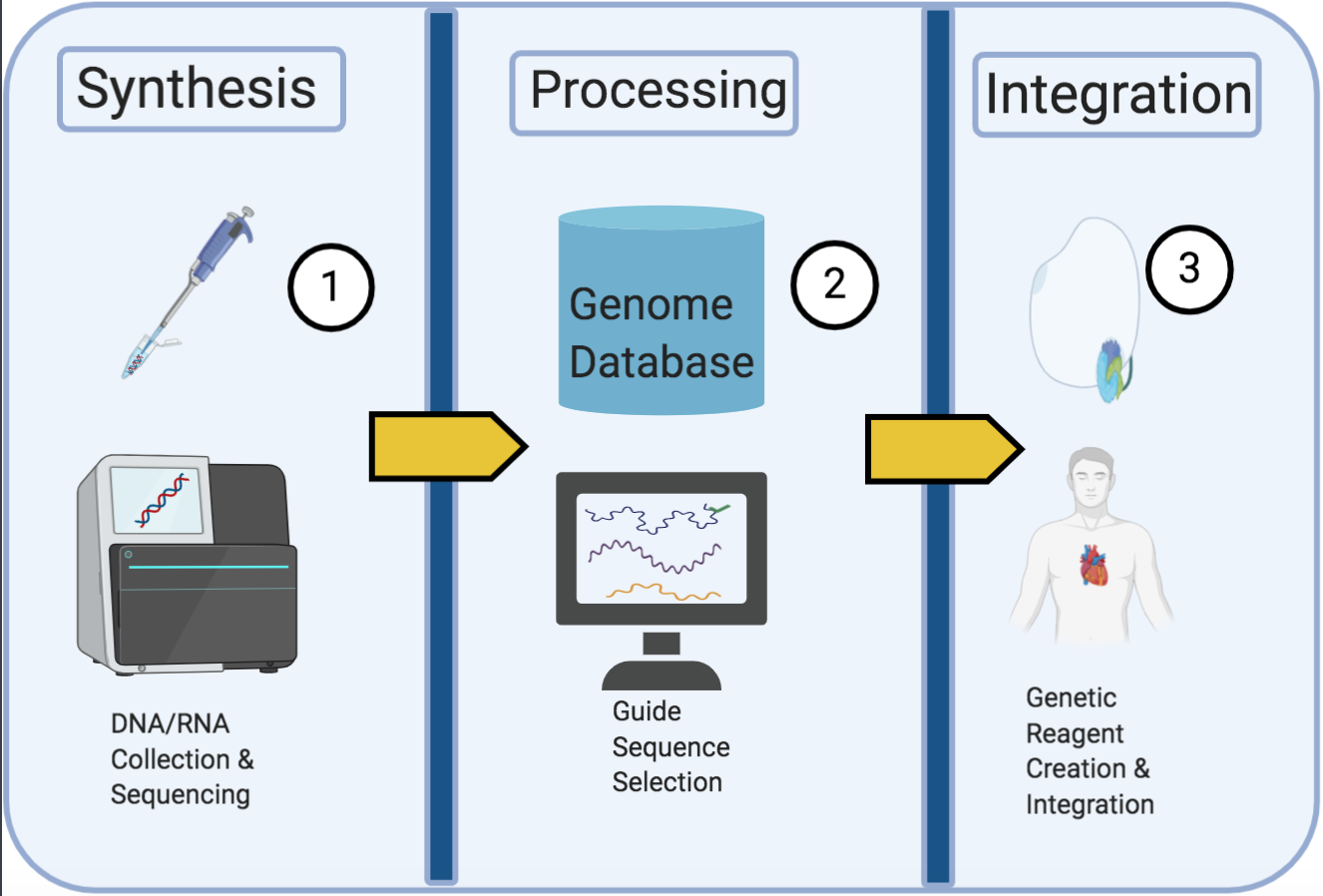}
  \caption{A general model for genetic modification. The process on the user end may start with the collection of the DNA or RNA (1), or pulling down the information from a database (2) if the sequence has already been cataloged. This solution protects the boundaries between (1) and (2) as well as (2) and (3).}
  \label{fig:gene editing model}
\end{wrapfigure}

The field of machine learning is growing and provides solutions to modern day problems such as facial recognition and fraud detection. Neural networks in particular, deep neural networks (DNN), are a key component of machine learning that can be adapted to understand a given domain. In this work, we apply a special kind of DNN called a convolutional neural network (CNN) to biological modeling, and show that machine learning can be used to predict and categorize in this domain.

In this paper, we enumerate targets in the current state-of-the-art of gene editing techniques, in which most of the components are implemented manually in labs, as well as a proposed implementation of machine learning to identify and remove risks. The contributions of this paper are: 
\begin{itemize}
    \item We outline finding in the CRISPR space, specifically aspects of the domain that demonstrate features of data set. 
    \item We present \textbf{Guide-Guard}, a convolutional neural network based solution for predicting the safety of a given guide sequence. 
    \item We outline challenges in the data-mining domain faced in creating \textbf{Guide-Guard}, as well as how we over came them. 
    \item We validate Guide-Guard's effectiveness on a real world CRISPR Cas13 database~\cite{Wessels}.
\end{itemize}


\section{Background and Related Work}
\label{sec:background-and-terminology}

\subsection{CRISPR}
\label{def:crispr}

\begin{wrapfigure}{i}{.5\textwidth}
  \centering
  \includegraphics[height=0.25\textwidth]{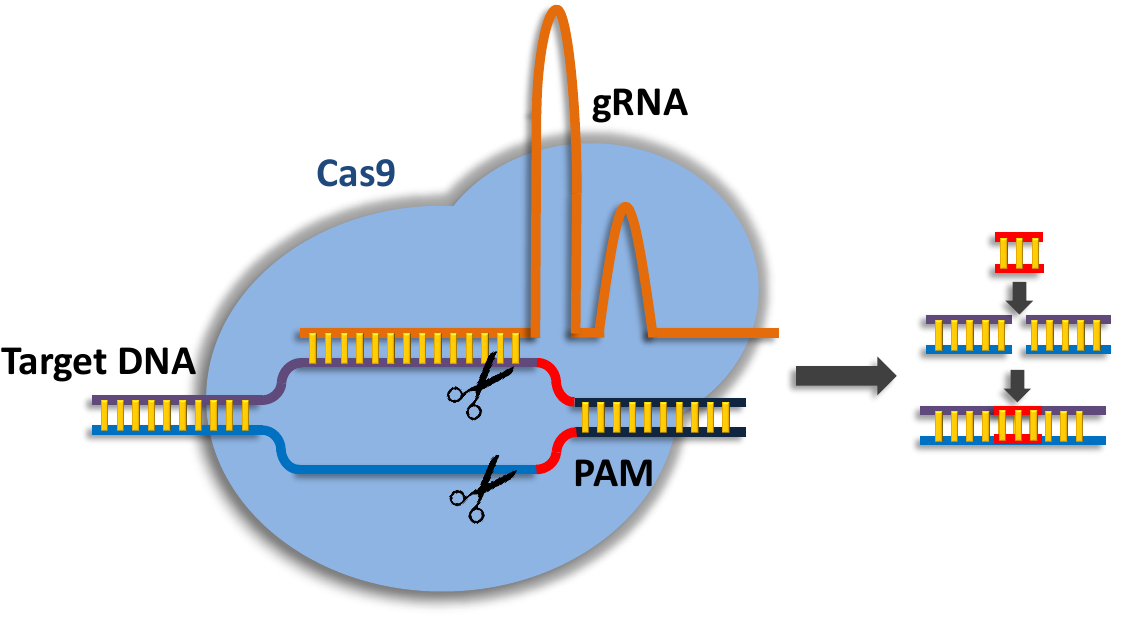}
  \caption{A pictorial representation of CRISPR Cas9 in action. Note, CRISPR Cas13 functions nearly identically, but on RNA instead.}
  \label{fig:cas9}
\end{wrapfigure}

\textbf{CRISPR} stands for Clustered Regularly Interspaced Short Palindromic Repeats, and comes in many varieties. The most famous and used are CRISPR Cas9\footnote{Cas stands for CRISPR-associated and is a protein required for DNA/RNA editing.} and Cas13~\cite{xu2020crispr} that are two proteins with different functionalities. This work focuses on the latter which modifies RNA in a similar fashion to Cas9 in figure Fig.~\ref{fig:cas9}. Another key difference is that Cas13 does not require a marker, Protospacer Adjacent Motif (\textit{PAM}), to index the regions that can be edited. Save for those, the process of using various kinds of CRISPR is the same.






\section{A GENERAL MODEL FOR GENE EDITING}
\label{sec:a-general-model-for-gene-editing}

Fig.~\ref{fig:gene editing model} shows a high-level view of the major stages in planning for and executing a gene editing experiment. The biologist begins with a target DNA/RNA being identified and sequenced~\cite{sequencing}. DNA/RNA sequencing is the process of determining the nucleic acid sequence and the order of nucleotides in DNA/RNA. It includes sensors used to determine the order of the four bases in a given DNA or RNA molecule. using a variety of cyber-physical sensing and processing platforms. If the sequence has already been processed, then the previous step may be skipped entirely. 

Target sequences are usually searched for off-target binding potential using web tools such as CRISPR Design Tool\footnote{\url{http://crispr.mit.edu}} or CCTop\footnote{\url{https://crispr.cos.uni-heidelberg.de}} and scored based on the potential for unintended binding. This information is forwarded onto the gene editing stage that includes obtaining synthesized primers and guide RNA (as synthesized sgRNA or as cr/tracr RNAs). The bench top protocols are performed in the lab, and the gene-edited target is sequenced and verified by comparison to the sequence. 


\section{Methodology}
\label{sec:methodology}
\begin{wrapfigure}{r}{.6\textwidth}
  \centering
  \includegraphics[height=0.35\textwidth]{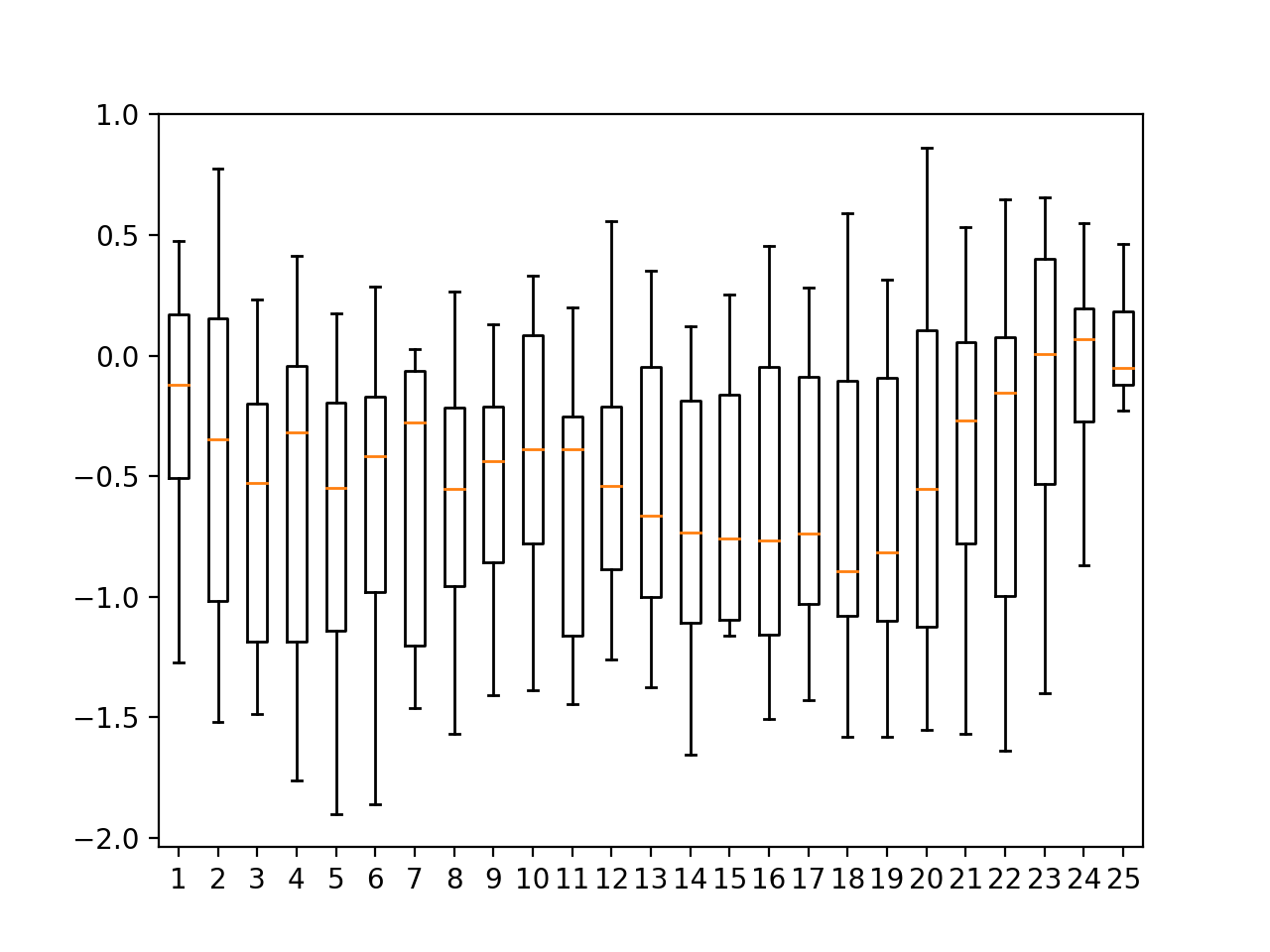}
  \caption{A histogram of the binding potential where three mismatches occurs next to each other. The index given if the first mismatch.}
  \label{fig:triple_mis}
  \vspace{-2em}
\end{wrapfigure}

For the following experiments, we used the data set from~\cite{Wessels}. We chose this data set because it included multiple different types of intentionally induced mismatches, which allows us to study what happens if an accidentally happens in practice. Also, it has multiple genes within it, allowing for a better understanding of the generalizability of the these findings. Lastly, it is a standard to benchmark our solution in an objective measure, given the best sequences are known. 

\subsection{Mismatch Location}

\begin{figure}
\centering
\parbox{5cm}{
\includegraphics[width=0.5\textwidth]{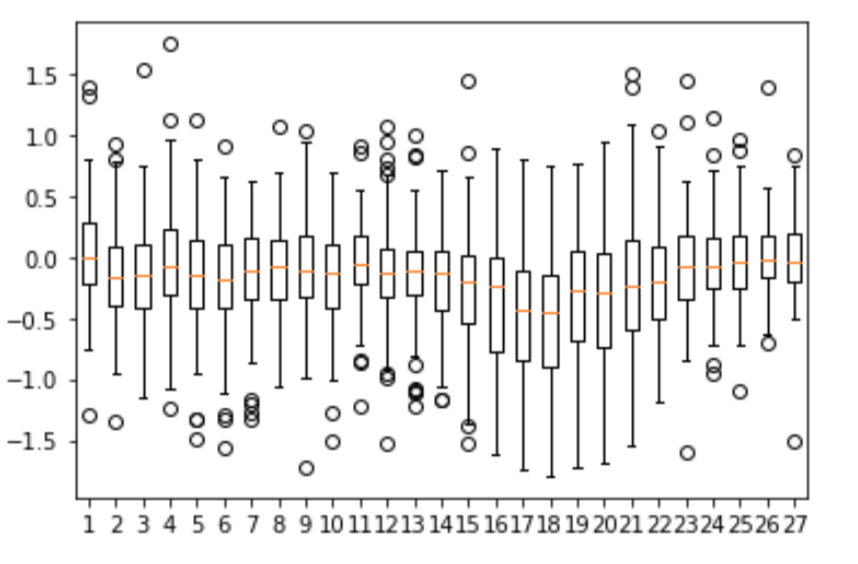}
\caption{A histogram detailing the binding potential as related to where a single mismatch occurs in the guide sequence.}
\label{fig:single_mis}}
\qquad
\parbox{5cm}{
\includegraphics[width=.5\textwidth]{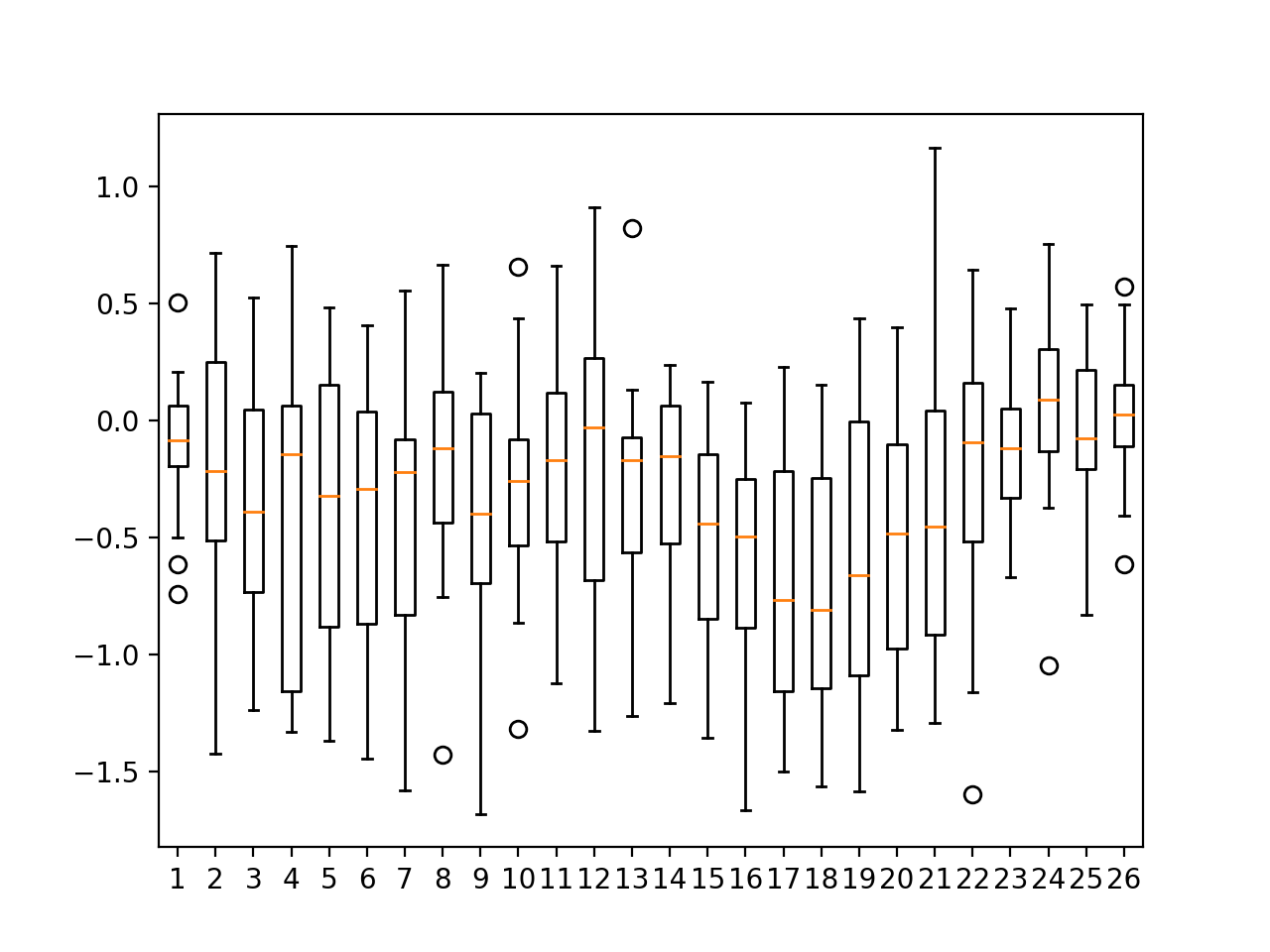}
\caption{The binding potential as related to where two mismatches occurs next to each other. The index is of the first mismatch.}
\label{fig:double_mis}}
\end{figure}

From the data, we observed that for a single mismatch, there is a clear impact based on the location Fig.~\ref{fig:single_mis}. From the histogram, we can see that the 18th nucleotide has a high importance in the CRISPR Cas13 binding energy. Additionally, the genes directly next to it also seem to have strong effect that tapers off the further away from the 18th nucleotide. This creates a normal distribution, with norm around the 18th. 

This trend is continued into the triple consecutive mismatch histogram Fig.~\ref{fig:triple_mis}. When there are three consecutive mismatches present in the guide sequence, the bimodal distribution is more clear. The peak at the 5th is more pronounced, and the peak at the 18th persists. Note as well that generally all of the binding energies decrease, which is what is expected with further deviation from the perfect match guide.

The previous three experiments have been of consecutive mismatches. In the next Fig.~\ref{fig:double_nc_mis}, we examine what happens when two different mismatches occur, but not necessarily consecutively as was the case in Fig.~\ref{fig:double_mis}. In this heat map, only the upper triangle is populated as the lower triangle would simply be a reflection. Also, the darker the cell is, the higher the activation energy. Note that the lowest activations are associated with the 5th and 18th nucleotides, consistent with our previous findings.   

\begin{wrapfigure}{i}{.55\textwidth}
  \centering
  \vspace{-2em}
  \includegraphics[height=0.34\textwidth]{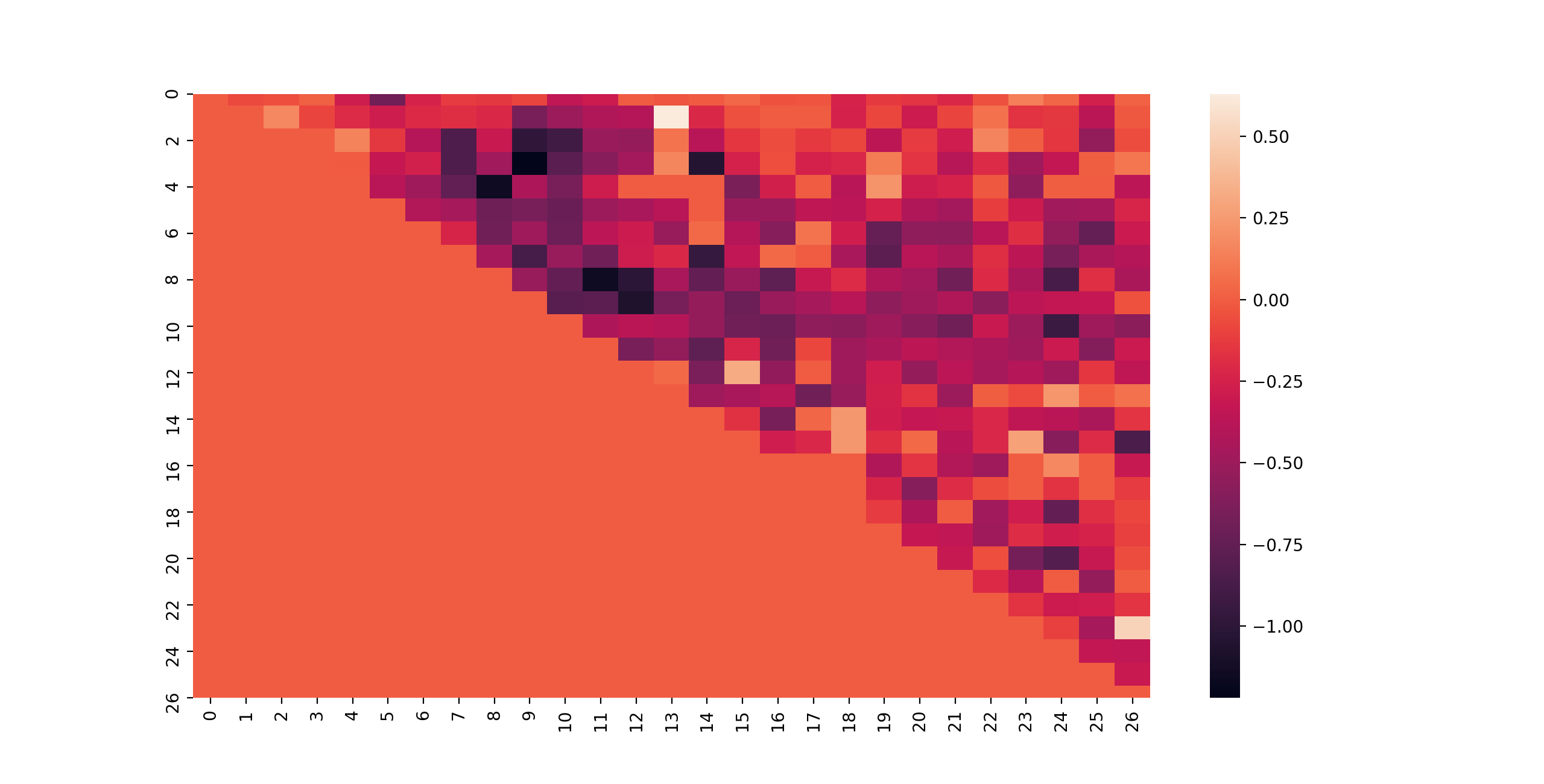}
  \caption{A heat map detailing the binding potential as related to where two mismatches occur within a guide sequence, but may not be next to each other. }
  \label{fig:double_nc_mis}
  \vspace{-3em}
\end{wrapfigure}

All of these values demonstrate the behavior of the guide sequence mismatch. These findings accord with the tertiary and secondary structure, which are the geometric structure of the gRNA. We used this information to put influence onto the channels associated with the 5th and 18th nucleotides. When we did this, we saw a \textbf{5.4\% boost to accuracy} compared to when we did not place emphasis.

\subsection{Nucleotide Replaced}

Another noteworthy aspect of the data is that the nucleotide replaced, that is the original nucleotide in the target sequence which is mismatched in the guide sequence.

\begin{figure}[t]
\centering
\parbox{5cm}{
\includegraphics[width=0.5\textwidth]{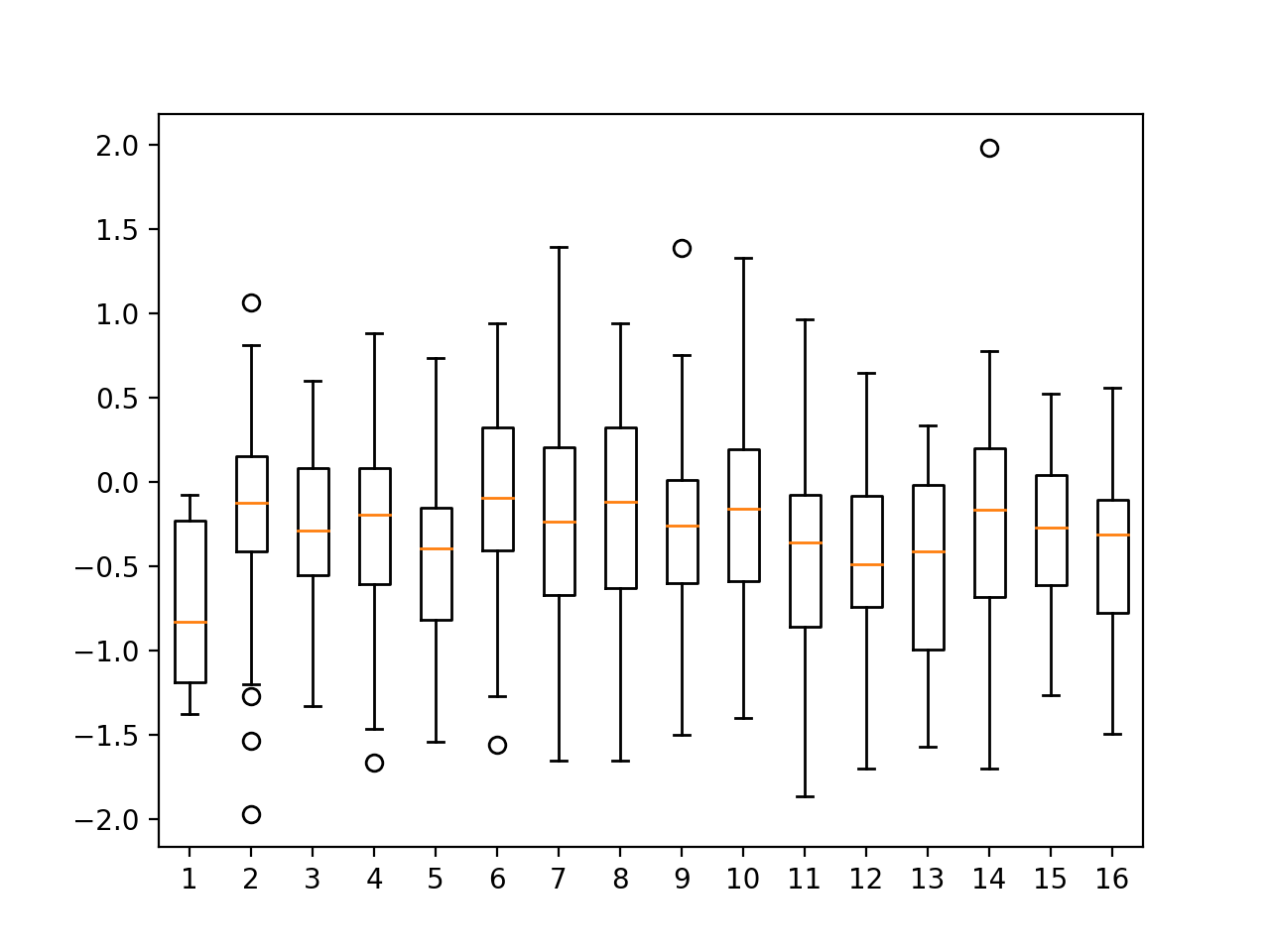}
\caption{A histogram detailing the binding potential as related to where a single mismatch occurs in the guide sequence, given it was originally an A.}
\label{fig:A_mis}}
\qquad
\parbox{5cm}{
\includegraphics[width=.5\textwidth]{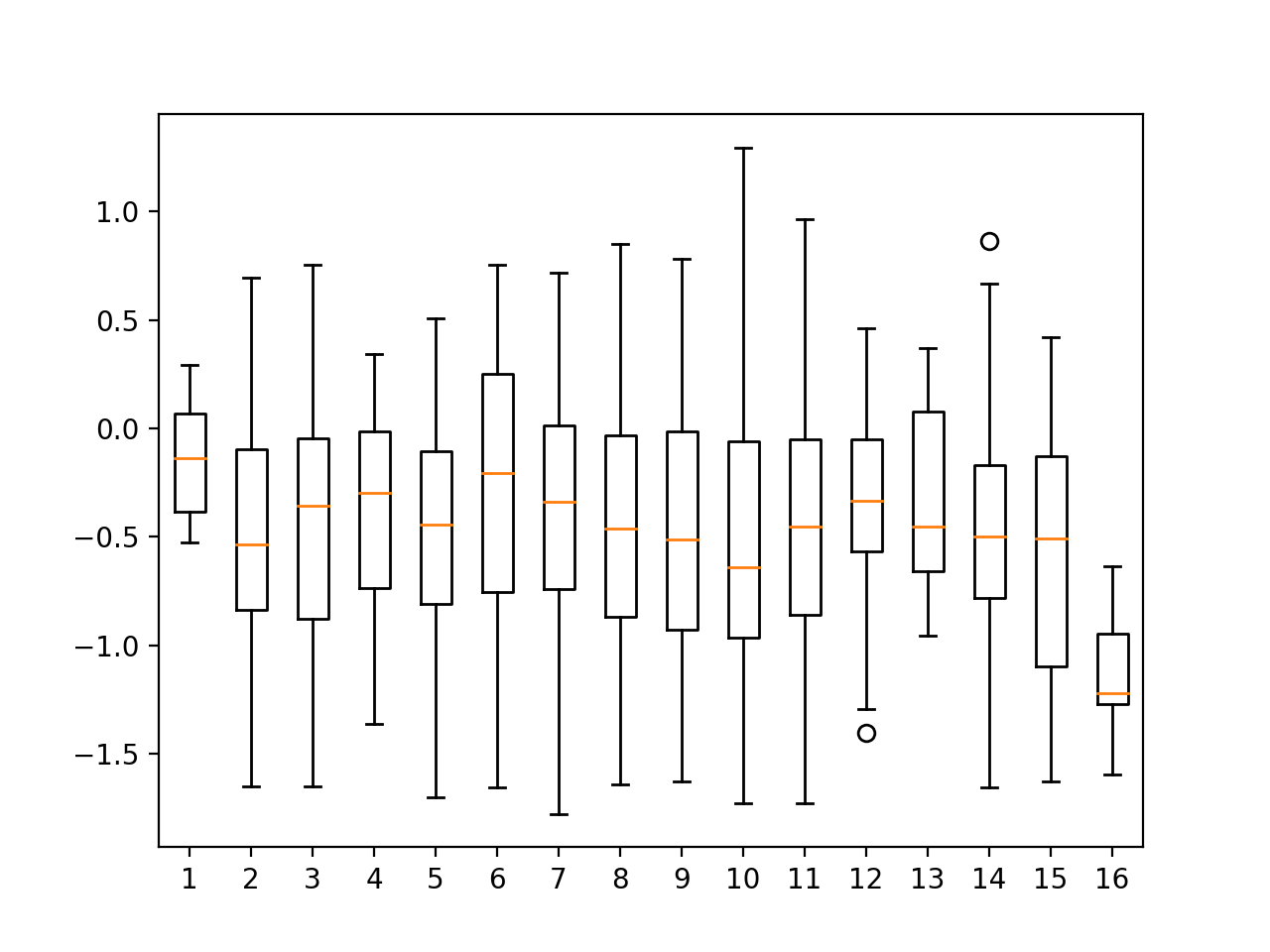}
\caption{A histogram detailing the binding potential as related to where a single mismatch occurs in the guide sequence, given it was originally an C.}
\label{fig:C_mis}}
\vspace{-2em}
\end{figure}

\begin{figure}[h]
\centering
\parbox{5cm}{
\includegraphics[width=0.5\textwidth]{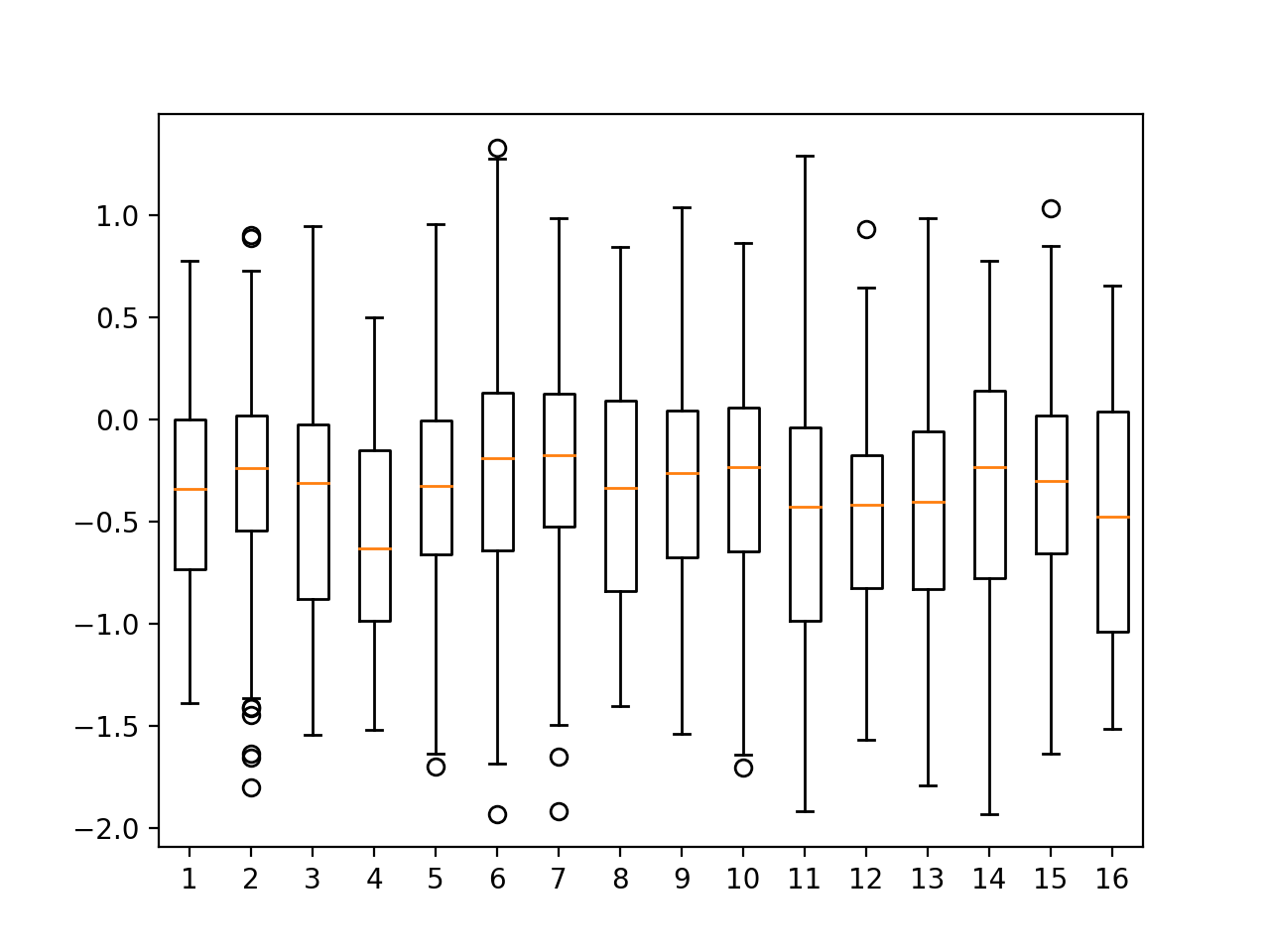}
\caption{The potential where a mismatch occurs in the guide sequence, given it was originally U. Although the dataset uses T, it should be U as Cas13 is for RNA.}
\label{fig:T_mis}}
\qquad
\parbox{5cm}{
\includegraphics[width=.5\textwidth]{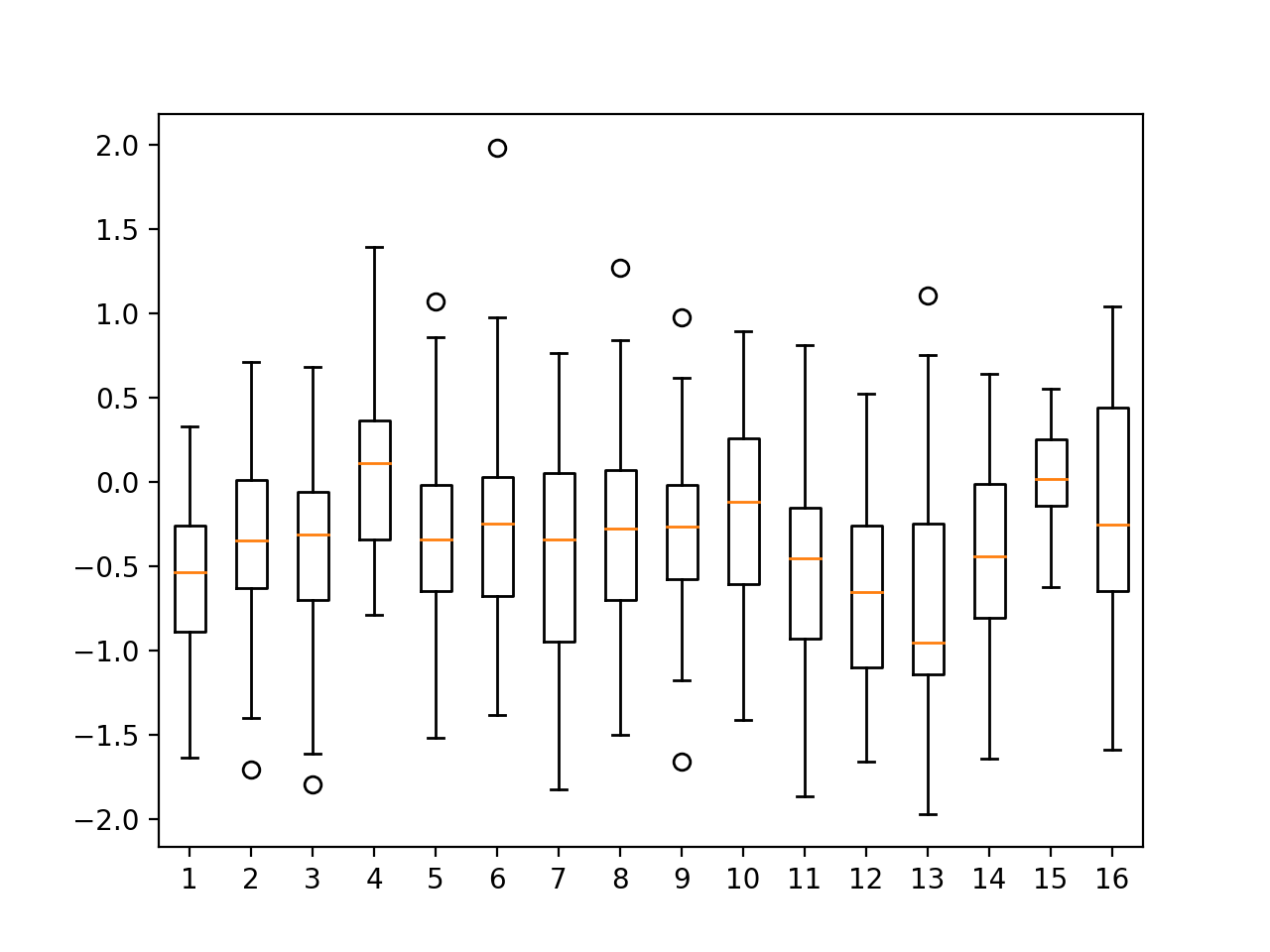}
\caption{A histogram detailing the binding potential as related to where a single mismatch occurs in the guide sequence, given it was originally an G. }
\label{fig:G_mis}}
\vspace{-2em}
\end{figure}

As can be seen from Fig.~\ref{fig:A_mis} Fig.~\ref{fig:C_mis} Fig.~\ref{fig:G_mis} Fig.~\ref{fig:T_mis} that U has the least effect when replaced, and G and C have higher effect. As such, in our set up of the model, we gave U a higher input strength to start with to provide the model with a deeper understanding of the space. When this was done, we saw a 3.8\% increase to accuracy when compared to fully even input values.

\subsection{Guide-Guard}

In order to increase the dependability from the malicious actors in either the synthesis phase to the processing phase (1 to 2 in Fig.~\ref{fig:gene editing model} and from the processing phase to the integration phase (1 to 3 in Fig.~\ref{fig:gene editing model}), or from user error, we designed Guide-Guard, using deep learning, to classify malicious gRNA that can be applied before use of gRNA. Wessels et al.~\cite{Wessels} compiled a data set  of RNA that can be analyzed to find their knockdown efficacy. The data set  is used for predictions of guide RNA on target RNA, which is the same problem as finding malicious guides. From this data set, we make use of 3 transcriptomes, CD46, CD55, and CD71. There are about 5,000 guides to apply to each of these 3 transcriptomes in separate libraries for each.

\begin{wrapfigure}{r}{.5\textwidth}
\vspace{-2em}
  \centering
  \includegraphics[height=0.27\textwidth]{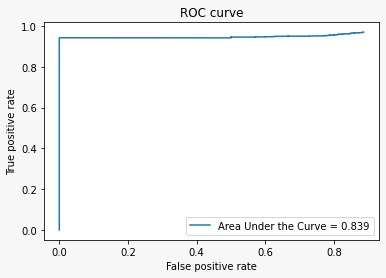}
  \caption{The ROC representing our results in correctly identifying malicious guide sequences.}
  \label{fig:results}
  \vspace{-3em}
\end{wrapfigure}

\subsection{Data Preparation}

We begin with feature extraction by one-hot encoding our nucleotides into a computationally comprehensible format for our network. Leveraging the results from the survey, we weigh the 18th and to a lesser extent 5th nucleotide and G/C's more within the encoding. This is done for both the proposed guide sequence as well as the reverse-complement of target sequence. These are then zipped together, as apposed to concatenated, so that there are 46 encoded nucleotides (23 from the guides and 23 from the targets) being presented to the model. When we tested concatenating the two sequences one after the other, we found that this out preformed zipping the sequences together when using a fully connected network, however when using a CNN, as we did here, we found that zipping them had an over all higher accuracy. This is a departure from previous works that assisted our results.

We then equally divide the data set by activation energy into 8. We chose 8 classes as this was the closes to having the edge of the top class be the cut off value for effective guide sequences found in~\cite{Metsky}. There are 8 classes we assign to the data set, with 1 class (the top ones having the highest activation energy) being for positive inputs and 7 of them representing negative inputs. The classes are evenly distributed by number of data points to have an unbiased data set  despite a large amount of negative inputs. We then train and validate using this data on our neural network, which is comprised of primarily dense layers and outputs one of these 8 classes. The highest of the 8 is a positive and secure input. The other 8 lack the knockdown efficacy to be safe for use. By having 8 equally distributed classes, we achieved the highest accuracy we could our of the tested division assignments as seen in Fig.~\ref{fig:results}.

\subsection{Network Design}

Guide-guard is a convolutional neural network, or CNN, based application. This was chosen due to CNN's ability to extract features from locality sensitive data, as such it makes a good choice for guide sequence selection. Its input layer takes in 46 values for the 23 nucleotides from each of the guide and target, with a kernel of 3 so the model is looking at a nucleotide and its neighbors. This is then put into another convolutional layer, which reduces the values down futher. These value are fed to a max pooling layer, which is finally flattened. These features are then piped to a fully connected network that goes from 400 to 200 to 100 to 50 to 25 and then finally 8 nodes. All of the activation functions utilized were ReLU, except the final layer, which utilizes softmax for classification. Since our goal was not to predict the exact activation energies, but rather to classify if a guide is sufficient for a given target, we utilized the categorical crossentropy loss function, and the Adam optimizer with a learning rate of .001.

\section{Results and Discussion}
\label{sec:results}

It should be noted that for our testing we accept non-perfect matches for guide sequences for non-malicious inputs. These results can be seen in table~\ref{tab:acc}. We rate according to knockdown efficacy and thus take into account these non-perfect matches, as those appear quite regularly without any malicious intent. This is why the neural network is needed, since it is not as easily distinguishable as perfect or non-perfect match. The knockdown efficacy overall must be taken into account in order to correctly classify the harmful and secure inputs.

\begin{table}[t]
    \centering
    
    \caption{This table takes a more fine grained look of guide-guard. These are the accuracies and performance measures of the solution on guide sequences that are perfect matches as well as off targets.}
    \begin{tabular}{ c | c | c }
\hline
                   & \textbf{Perfect Matches} & \textbf{Mismatch} \\ \hline \hline
\textbf{Accuracy}           & 85.51\%         & 77.50\%    \\ \hline
\textbf{True Positive Rate} & 98.87\%         & 98.44\%    \\ \hline
\textbf{True Negative Rate} & 79.48\%         & 66.92\%    \\ \hline
    \end{tabular}
    \label{tab:acc}
\end{table}

Our solution correctly classified about 84\% of the inputs when tested against this data. To do this we use a 20-fold cross validation of our network on the data set  to ensure proper results. Fig.~\ref{fig:results} represents the performance of our deep neural network at different classification thresholds. The receiver operating curve shows our true positive rate in comparison to our true negative rate with the rating being the area under the curve. Our area under the curve was 0.839 as seen in Fig.~\ref{fig:results}, which is quite high. Our security checkpoint, while accurate, is also computationally efficient with a runtime of only 0.00055 seconds on average for a singular input. This average was found when run on 1 fold, or 1/20th, of the total data set.

As can be seen in table~\ref{tab:acc}, guide-guard has a higher accuracy on perfect matches than on mismatch guides. Perfect matches are guides which are the exact same sequence as the target being changed. Mismatch guides, on the other hand, are guide sequences which differ slightly from the target in some way. This discrepancy in accuracies could be explained with the observation that mismatches have a much higher variation in values, some differences in guides having no effect to binding potential, while others having dramatically different values. This would make it harder to classify. Even with this, guide-guard does better than the current best methods, which only look at perfect matches. 


Our proposed solution can be used for anyone who needs to clarify the target sequence. This should be implemented before submitting to a database to secure against poisoning or any sort of errors that may occur.

Additionally, users should make use of this before applying RNA they have not developed themselves. This will add an extra layer of trust to the guide. With these implementations before Processing and Integration, as seen in Fig.~\ref{fig:gene editing model}, users can have a more secure and dependable experience.

This also applies to an automated approach with relative ease as well as the time required to verify a sample is only 0.00055 seconds on average when ran on a 2011 Macbook Pro 13". This means that even in higher scaled cases, the time required would increase by only about 5.5 seconds for every 10,000 inputs. When considering the potential dangers, or even just the potential time delays, from a single dangerous input, this would be remedied in 5.5 seconds.

When dangerous input warnings would appear as a result of our security program, the inputs can be checked and verified manually or discarded immediately. This allows automated and unverified processes to run smoother and more efficiently, along with adding security to any individual's CRISPR use. Databases and experiments can be secured through the implementation of our screening process before inputs are made on either one, steps 2 and 3 of Fig.~\ref{fig:gene editing model}.

\section{Conclusion}
\label{sec:conclusion}

The almost uninhibited accessibility to CRISPR has facilitated a new era of genome editing that is rapidly increasing the ceiling of innovation in biological design. However, the accessibility to such powerful technology must be done in a trustworthy way to mitigate threats of off-target behavior. In this paper, we propose Guide-guard as a solution to identify poor guides to a CRISPR Cas13 user's work, which can be used to stop any serious dangers to the end users.
%
%
%
\bibliographystyle{splncs04}
\bibliography{samplepaper}

@article{xu2020crispr,
	Author = {Xu, Dongyang and Cai, Ye and Tang, Lu and Han, Xueer and Gao, Fan and Cao, Huifen and Qi, Fei and Kapranov, Philipp},
	Da = {2020/02/04},
	Doi = {10.1038/s41598-020-58104-5},
	Id = {Xu2020},
	Isbn = {2045-2322},
	Journal = {Scientific Reports},
	Number = {1},
	Pages = {1794},
	Title = {A CRISPR/Cas13-based approach demonstrates biological relevance of vlinc class of long non-coding RNAs in anticancer drug response},
	Ty = {JOUR},
	Url = {https://doi.org/10.1038/s41598-020-58104-5},
	Volume = {10},
	Year = {2020}}

@article {Metsky,
	author = {Metsky, Hayden C. and Welch, Nicole L. and Haradhvala, Nicholas J. and Rumker, Laurie and Zhang, Yibin B. and Pillai, Priya P. and Yang, David K. and Ackerman, Cheri M. and Weller, Juliane and Blainey, Paul C. and Myhrvold, Cameron and Mitzenmacher, Michael and Sabeti, Pardis C.},
	title = {Efficient design of maximally active and specific nucleic acid diagnostics for thousands of viruses},
	elocation-id = {2020.11.28.401877},
	year = {2020},
	doi = {10.1101/2020.11.28.401877},
	publisher = {Cold Spring Harbor Laboratory},
	URL = {https://www.biorxiv.org/content/early/2020/11/28/2020.11.28.401877},
	eprint = {https://www.biorxiv.org/content/early/2020/11/28/2020.11.28.401877.full.pdf},
	journal = {bioRxiv}
}

@article{Wessels,
    author = {Wessels, Hans-Hermann and Méndez-Mancilla, Alejandro and Guo, Xinyi and Legut, Mateusz and Daniloski, Zharko and Sanjana, Neville},
    year = {2020},
    month = {06},
    pages = {},
    title = {Massively parallel Cas13 screens reveal principles for guide RNA design},
    volume = {38},
    journal = {Nature Biotechnology},
    doi = {10.1038/s41587-020-0456-9}
}

@article{sequencing,
	An = {26554401},
	Author = {Heather, James M and Chain, Benjamin},
	Date = {2016/01/},
	Db = {PubMed},
	Doi = {10.1016/j.ygeno.2015.11.003},
	Et = {2015/11/10},
	Isbn = {1089-8646; 0888-7543},
	J2 = {Genomics},
	Journal = {Genomics},
	L2 = {https://www.ncbi.nlm.nih.gov/pmc/articles/PMC4727787/},
	La = {eng},
	Month = {01},
	Number = {1},
	Pages = {1--8},
	Publisher = {Academic Press},
	Title = {The sequence of sequencers: The history of sequencing DNA},
	Ty = {JOUR},
	U1 = {26554401{$[$}pmid{$]$}},
	U2 = {PMC4727787{$[$}pmcid{$]$}},
	U4 = {S0888-7543(15)30041-0{$[$}PII{$]$}},
	Url = {https://pubmed.ncbi.nlm.nih.gov/26554401},
	Volume = {107},
	Year = {2016}}

@article{testing,
	Author = {You, Yan and Ramachandra, Sharmila G. and Jin, Tian},
	Da = {2020/09/08},
	Doi = {10.1038/s41598-020-71690-8},
	Id = {You2020},
	Isbn = {2045-2322},
	Journal = {Scientific Reports},
	Number = {1},
	Pages = {14779},
	Title = {A CRISPR-based method for testing the essentiality of a gene},
	Ty = {JOUR},
	Url = {https://doi.org/10.1038/s41598-020-71690-8},
	Volume = {10},
	Year = {2020}}

@article{guide_seq,
	Author = {Wang, Daqi and Zhang, Chengdong and Wang, Bei and Li, Bin and Wang, Qiang and Liu, Dong and Wang, Hongyan and Zhou, Yan and Shi, Leming and Lan, Feng and Wang, Yongming},
	Da = {2019/09/19},
	Doi = {10.1038/s41467-019-12281-8},
	Id = {Wang2019},
	Isbn = {2041-1723},
	Journal = {Nature Communications},
	Number = {1},
	Pages = {4284},
	Title = {Optimized CRISPR guide RNA design for two high-fidelity Cas9 variants by deep learning},
	Ty = {JOUR},
	Url = {https://doi.org/10.1038/s41467-019-12281-8},
	Volume = {10},
	Year = {2019}}

@article{RATH2015119,
title = {The CRISPR-Cas immune system: Biology, mechanisms and applications},
journal = {Biochimie},
volume = {117},
pages = {119-128},
year = {2015},
note = {Special Issue: Regulatory RNAs},
issn = {0300-9084},
doi = {https://doi.org/10.1016/j.biochi.2015.03.025},
url = {https://www.sciencedirect.com/science/article/pii/S0300908415001042},
author = {Devashish Rath and Lina Amlinger and Archana Rath and Magnus Lundgren}
}

@article{transcriptome,
	An = {22916334},
	Author = {Pertea, Mihaela},
	Date = {2012/09/},
	Db = {PubMed},
	Doi = {10.3390/genes3030344},
	Isbn = {2073-4425; 2073-4425},
	J2 = {Genes (Basel)},
	Journal = {Genes},
    L2 = {https://www.ncbi.nlm.nih.gov/pmc/articles/PMC3422666/},
	La = {eng},
	Month = {09},
	Number = {3},
	Pages = {344--360},
	Publisher = {MDPI},
	Title = {The human transcriptome: an unfinished story},
	Ty = {JOUR},
	U1 = {22916334{$[$}pmid{$]$}},
	U2 = {PMC3422666{$[$}pmcid{$]$}},
	Url = {https://pubmed.ncbi.nlm.nih.gov/22916334},
	Volume = {3},
	Year = {2012}}

\end{document}